# Dense Adaptive Cascade Forest: A Self Adaptive Deep Ensemble for Classification Problems

Haiyang Wang [1]
Southwest Jiaotong University
Chengdu, China
why529570509@163.com
ORCID: 0000-0003-2273-2686

Yong Tang [2]
University of Electronic Science and Technology of China
Chengdu, China
tangyong@uestc.edu.cn
ORCID: 0000-0002-5036-4350

Ziyang Jia [2]
Rutgers University
New Jersey, USA
ziyang.jia@rutgers.edu
ORCID: 0000-0001-6686-0845

Fei Ye [3]
Southwest Jiaotong University
Chengdu, China
122404504@qq.com
ORCID: 0000-0002-5894-2178

**Abstract.** Recent researches have shown that deep forest ensemble achieves a considerable increase in classification accuracy compared with the general ensemble learning methods, especially when the training set is small. In this paper, we take advantage of deep forest ensemble and introduce the Dense Adaptive Cascade Forest (daForest). Our model has a better performance than the original Cascade Forest with three major features: first, we apply SAMME.R boosting algorithm to improve the performance of the model. It guarantees the improvement as the number of layers increases. Second, our model connects each layer to the subsequent ones in a feed-forward fashion, which enhances the capability of the model to resist performance degeneration. Third, we add a hyper-parameters optimization layer before the first classification layer, making our model spend less time to set up and find the optimal hyper-parameters. Experimental results show that daForest performs significantly well, and in some cases, even outperforms neural networks and achieves state-of-the-art results.

## 1 Introduction

Recent years have witnessed a phenomenal emergence of deep neural networks (DNN) [1, 2]. In many fields, such as image classification and speech recognition, DNNs even become the dominant approach [4]. With the improvement of computer hardware, researchers have the ability to train increasingly deeper and more complex networks. Eventually, neural networks gradually surpassed humans in solving specific problems [8, 9, 10].

For DNNs mentioned above, we generally refer to convolutional neural networks, which are composed of convolutional layers and fully-connected layers [2]. Convolutional layers assist models to extract high-order features from raw data and feed these features into fully-connected layers to get the final prediction result. Many experiments have proved that this mechanism is efficient and useful [5, 6]. However, as a classifier, fully-connected layers did not change much ever since they were invented and used almost 30 years ago [5, 7]. Moreover, it is widely recognized that fully-connected network is highly prone to overfitting when the size of training data is relatively small. Imagine that the training set is smaller than the number of neural network parameters, and with extremely high possibility, the network will converge to local optimums and overfit. Dropout and regularization are widely used to alleviate this problem [11, 12, 13]. However, overfitting is a severe problem which is hard to avoid and solve thoroughly for neural networks. As DNNs gradually become deeper, more powerful and more efficient, unfortunately more new research problems emerge:

– Even in the field of big data, a large portion of the data is not spatial or temporal correlated, leading to the fact that most DNNs cannot be applied directly. To fully exploit the large training sets, researchers have to utilize ordinary learning models [23, 30, 48], such as linear or generalized linear models, support vector machine, random forest, etc. Compared with DNNs, these models are structurally simple and relatively less inclined to overfit [15, 45, 47].

– In many circumstances, collecting massive data is impossible or costly. Numerous studies have shown that transfer learning [16] can make DNN work well on new datasets[17, 18]. However, since it cannot avoid the problems of data size or even the uncorrelation on spatial and temporal, transfer learning is not the key. As for traditional learning models, small-scale data can fully train these ordinary models, and ensemble can significantly improve the capacity of such models [15, 20].

– The biggest problem of DNN is the non-interpretability, which makes it extremely difficult to apply theoretical analysis on DNNs, leading it to be treated as a black box [19]. Compared with DNN, traditional learning models [50, 23, 24] have better interpretability. It is easier to apply theoretical analysis to these models, although such analysis is beyond the scope of this paper.

The model we proposed has the characteristics of both ensemble and deep. In order to alleviate the deficiencies mentioned above, we propose a novel decision tree ensemble method named *dense adaptive cascade forest* (daForest) which fully utilizes the potential of the ensemble. Our experiments show that ensemble models can enhance the representational ability of individual models and achieve better performance on prediction and generalization [15, 20]. Motivated by the state-of-the-art results achieved by several recent DNN approaches [21, 22], the depth of network is one of the key factors that can contribute to good performance. As the models go deeper, the latter layers capture features of higher-order.

Besides, we notice that the degree of discrimination is different among samples, suggesting the contribution of different samples in the model training process is not the same. However, original stacking model or deep model does not make good use of this fact [15, 20, 14], so we introduce an adaptive boosting approach to enhance performance of our model. Adaptive boosting can adjust weights of samples and turn the whole cascade model into a deep additive model [25, 26, 27, 28]. Such transformation endows some particularly good features into a deep cascade model. daForest can distinguish different samples with different degree of discrimination. Thus, the model can adjust its attention to more important sample in each layer. More importantly, the final prediction can make full use of the superior predictions of each layer because of the weighted additive characteristic of adaptive boosting. Deep ensemble and boosting make daForest not just stack base classifier to obtain a large nominal depth, but fully release the potential of the deep model. In daForest, we also employ a distinctive architecture called dense connectivity, which is first introduced in DenseNet [29]. The main contributions of our work are as follows:

(1) We apply a layer-wise boosting procedure to the deep forest stacking model [14], which can improve the performance of such stacking structures significantly without too much burden of calculation. To the best of our knowledge, we are the first to combine layer-wise boosting method with deep stacking ensemble model. Studies found that ensemble approaches enable the models to achieve a reasonable balance between bias and variance [15, 45, 47].

(2) daForest is a deep model, which means that information sharing between each layer is essential in training and inferencing stages. The distribution of samples can be changed in each layer by the concatenation of raw features and newly generated probabilistic features, but not always in the expected direction. However, information sharing gives each layer a chance to correct its mistakes. Inspired by DenseNet [29], we introduce dense connectivity to the proposed model, which ensures maximum information flows between the layers.

(3) We add a hyper-parameters optimization layer before the first classification layer. The proposed deep model also suffers from overfitting and calculation burden, the same as DNNs. However, in our approach, the basic unit is forest [23, 24], the ensemble of decision trees, and the capacity and computation scale of each forest are determined, to a large extent, by the size of tree ensemble. In order to solve such problems, we propose an architecture that distills this insight into to a simple linear search method. This ensures our model can achieve a balance between capacity and computation scale.

In the paper, we compare our model with several machine learning models, including convolutional neural networks, dimension reduction methods and ordinary learning models, such as MLP, SVM, Random Forest and Logistic Regression, etc. Most importantly, we also compare our model with gcForest, which is first introduced in [Zhou, 2017]. We found that our model achieves better performances than these learning models on 12 benchmark datasets, including both high-dimensional datasets (5000-10000 attributes) and low-dimensional datasets. The merits of the model are experimentally verified. A significance test is carried out to determine whether there are significant differences in the results of the classification algorithms. Besides the benefits on results, we also find that our model has a powerful ability to handle high-dimensional sparse data without any dimensional reduction preprocessing on the raw data, like PCA, projection pursuit, etc.

The rest of this paper is organized as follows: Section 2 reviews the related works in cascade forest and daForest. In section 3, we present the overall structure of daForest in detail. In section 4, we compare daForest with several popular machine learning models by giving experimental results and evaluations. Finally, in section 5, we conclude the paper with a summary of the comparison of daForest and gcForest followed by suggestions for the future research.

## 2    Related Work

daForest is an improved tree stacking model, an individual learner in daForest is an ensemble of random forest and extremely randomized trees [23, 24]. Forest is widely used [46, 49] and have a strong ability to resist overfitting [51]. Stacking, a widely used ensemble approach, usually takes the output of the preceding layer as the input of the next layer. Deep convolutional neural networks [2] and deep belief nets [34] are the most popular stacking models that opens a door on the eve of deep learning sweeping across the globe. Experimental and theoretical studies have proved that the stacking can indeed improve the performance of individual learners [15, 20]. Unlike deep neural network, composed of thousands of nonlinear differentiable base classifiers, which can be trained by backpropagation algorithm, the stacking method of the ordinary base learner is trickier than it looks. Without backpropagation, end to end training is impossible to be applied to ordinary stacking model, which severely restricts the depth of the stacking model and makes it more prone to overfitting. To relieve the problem to the maximum extent, we employ two methods: concatenating the output probabilistic features with original input features of the first layer in each subsequent layers (we call such stacking model with individual tree learner as cascade forest) and changing the weights of each sample through boosting [25, 26, 27, 28]. The proposed stacking approach is highly competitive with current ensemble methods (such as bagging, boosting and plain stacking) and is extremely easy to implement.

In 2017, Zhi-Hua Zhou et al. published their pioneering study on deep ensemble/stacking model named gcForest [14], which is a cascade forest facilitated with multi-grained scanning (working like the convolutional kernels) to extract spatial or temporal correlations within the features. The performances recorded are quite promising and inspiring: on minist and cifar-10 image classification datasets, gcForest outperforms deep belief nets and shallow convolutional neural networks [14]. Besides, on most of the discrete features dataset, such as IMDB, YEAST, ADULT, etc., gcForest even achieves state-of-the-art results. We have designed several experiments to compare daForest with gcForest and other classification methods. The results show that daForest has a better performance than other models, including gcForest, on these datasets. Compared with DNNs, gcForest has fewer hyper parameters and faster training speed. In addition, the structural complexity of the model is dynamically determined in the training process. Cascade forest can be regarded, to some extent, as a kind of boosting method, since it is a densely connected cascade forest boosted by SAMME.R algorithm [28], which makes daForest an enhanced boosting method.

As a key part of gcForest, cascade forest is firstly proposed in [Zhi-Hua Zhou et al. 2017]. Cascade forest is composed of multiple cascade levels, where each level of cascade layer takes input from the previous layer and outputs concatenation of probabilistic features and raw features as input of next layer. Each layer of cascade is an ensemble of forests, random forests and completely random forests are used as base classifiers in each cascade layer. As expounded in [Zhou 2012], diversity is a crucial part of ensemble constructions [14, 15], which can impede the performance of ensemble models greatly [14]. In cascade forest, researchers usually employ two methods to encourage diversity, the first is packaging together several random forests and completely random forests, and the second is using k-fold cross-validation to train each forest and generate augmented features. It is noteworthy that completely random forest is a variant of extremely random forest [24] and the main difference is the number of candidate features used in each tree node splitting. Besides, we can get a completely random forest by randomly selecting only one feature for splitting at each node of extremely random forest. To solve image and audio classification problems, gcForest employs a feature extraction approach named multi-grained scanning [14], acting similarly with the convolutional layers [1, 2] of DNN. Multi-grained scanning layers use several sliding windows of different sizes to scan the raw features, then cut them into small pieces. This process improves the performance of cascade forests by retaining the spatial and sequential relationships.

The cascade structure can prevent traditional stacking models [15, 20] from overfitting by employing raw feature combination, completely random forest [24] and k-fold cross-validation. However, overfitting and performance degeneration are still severe problems in cascade forests. Since that, as the number of layers goes bigger, the accuracy of model climbs a little higher in the first few layers then drops rapidly. While, what makes it worse, the feature diversity will be reduced in the subsequent training. To handle the problems above, in the proposed model, we design a mechanism of sample-wise attention and layer-wise information sharing to encourage feature diversity, which can effectively alleviate overfitting and performance degeneration. Besides, the number of trees in different kinds of forests is fixed across different tasks in cascade forest, which is not a wise choice, because the complexity of different classification tasks is variant. We introduce an optimization layer to find the best hyper-parameter of different types of forests to achieve a balance between task complexity and model capacity.

AdaBoost [25, 26] is a popular boosting algorithm, which can change the distribution of samples by assigning different weight to each of them. The weights of samples reflect the difficulty of classifying, which leads to the fact that the greater a sample weights, the higher the probability of being misclassified. AdaBoost is also an additive boosting method and each model, generated through the training process, has a different weight that reflects the performance of it. In the process of classification, some classifiers may have better results than the others on some indistinguishable samples. In this way, a complementary classifier series is established. SAMME.R is an improvement of original Adaboost, which naturally extends the original AdaBoost algorithm to the multi-class case without reducing it to multiple two-class problems [28] and reduces the requirement of minimum accuracy for individual classifiers. A study shows that AdaBoost can even be used to reduce the dimensionality of image features [31]. Experimental results in our work indicate that this algorithm can significantly enhance our deep model.

Our model employs random forest [23] and extremely random forest [24] as individual learners. By using bagging and randomly splitting features space, the random forest can achieve very good classification results and outperforms DNNs on non-spatial and non-temporal dataset. Extremely random forest, which randomly chooses subspace division in the construction of decision trees, is more random than ordinary random forest, which chooses the best division since it is well known that diversity is a pivotal characteristic to the ensemble models [15, 20]. Some studies shown that random forest can even be combined with DNNs [32, 33] and back propagate gradients through neurons.

## 3  The Proposed Deep Ensemble Approach

Fig. 1 shows the overall procedure of daForest. daForest consists of cascade forest [14], boosting procedure [28, 31], dense connectivity [29] and comprises $\mathcal{L}$ layers. Given input $\mathcal{X} \subset \mathcal{R}^d$ and output space $\mathcal{Y}$, each layer can be regarded as an individual ensemble module $E_l$, where $l$ indicates the index of layers, each of which implements a classification function $E_l(\cdot): \mathcal{X} \to [0, 1]$. An individual ensemble module has several individual learners $F_{li}$, which can be random forest [23] or extremely random forest [24] and apply transformation $F_{li}(\cdot): \mathcal{X} \to [0, 1]$ to each sample $x_i$. Finally, we get the predictive probability of $x_i$: $F_{li}(x_i)$. We indicate the original input of the first layer as $x_0$ and the output probabilistic features of $l^{th}$ layer as $x_l$. In this section, we will first introduce the structure of cascade forest, and then boosting procedure, followed by dense connectivity and optimization layer of daForest.

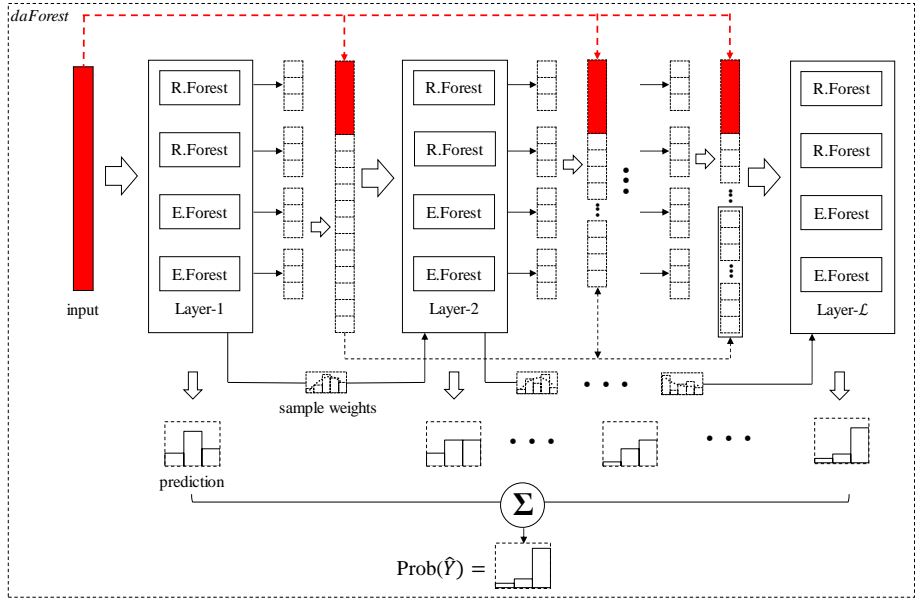

**Fig. 1.** The overall procedure of daForest. Each layer consists of 4 classifiers (2 random forests and 2 extremely random forests) distinguishing among 3 classes. For brevity, the hyper-parameter optimization layer has been omitted.

### 3.1  Cascade Forest

Cascade forest, first proposed in [Zhou, 2017], is the fundamental framework of the entire model. Fig. 2 illustrates the layout of a cascade forest. Cascade forest can be regarded as a deep ensemble of random forests and extremely random forests. By treating those two kinds of random forests as individual learners, we package them, for each layer, into a single ensemble module. Each layer of our cascade forest is composed of a single ensemble module, with the concatenation of original input features and the output probabilistic features of the previous layer as the input of the layer. It is noteworthy that concatenating probabilistic features and original input features into a single input feature vector is an effective method to prevent overfitting. Traditional stacking models connect the $l^{th}$ layer's output as the input to the $(l+1)^{th}$ layer, which can be described as:

$$x_l = E_l(x_{l-1}) = [F_{l1}(x_{l-1}), F_{l2}(x_{l-1}), \ldots, F_{ln}(x_{l-1})], l = 1, \ldots, \mathcal{L} \quad (1)$$

where $x_l$ is the probabilistic features produced by forests. However, such structure is highly prone to overfitting when the number of layers increases and may impede the diversity of individual learners, because each layer is only related to its preceding layer leading to the result that the individual learners will tend to be homogenized and the diversity of traditional stacking model is going to be worse when it's getting deeper. One way to solve the problem is combining original features $x_0$ and probabilistic features $x_l$ together before training $l^{th}$ layer's individual learners (as illustrated in Fig. 2):

$$x_l = [x_0, E_{l-1}(x_{l-1})] = [x_0, F_{l1}(x_{l-1}), \ldots, F_{ln}(x_{l-1})], l = 2, \ldots, \mathcal{L} \quad (2\text{-}1)$$
$$x_l = E_l(x_0) = [F_{l1}(x_0), \ldots, F_{ln}(x_0)], when\ l = 1 \quad (2\text{-}2)$$

and the final prediction of cascade forest is the output of the last layer. In order to get probabilistic features of each instance, we must record the class distribution at leaf nodes in each decision tree of the random forest. Followed with averaging the class percentage of instance we concern across all trees and concatenating probabilistic features generated by each forest in the same layer. We employ stratified k-fold validation in each layer so as to reduce the impact of overfitting and keep class distribution unchanged in each fold. For example, suppose that we use 3-fold cross validation with 2 forests in each layer, we cut training data X into 3 pieces that keep the same class distribution, denoted as [$X_1$, $X_2$, $X_3$], then feed the data to classifiers in three times and eventually get the prediction [$Y_1$, $Y_2$, $Y_3$]:

fold 1: training classifier $C_{11}$ and $C_{12}$ with [$X_2$, $X_3$], Y1=[$C_{11}(X_1)$, $C_{12}(X_1)$]
fold 2: training classifier $C_{21}$ and $C_{22}$ with [$X_1$, $X_3$], Y2=[$C_{21}(X_2)$, $C_{22}(X_2)$]
fold 3: training classifier $C_{31}$ and $C_{32}$ with [$X_1$, $X_2$], Y3=[$C_{31}(X_3)$, $C_{32}(X_3)$]

where the concatenation of [$Y_1$, $Y_2$, $Y_3$], noted as Y, will be passed to the next layer as input.

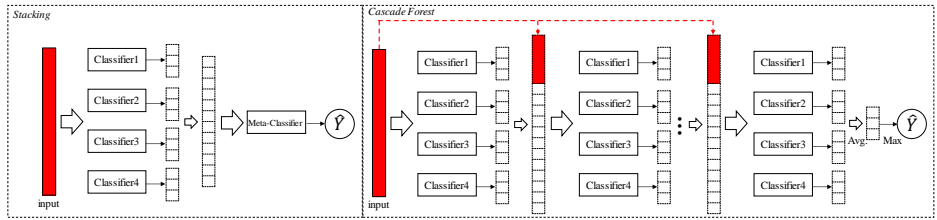

**Fig. 2.** left: traditional stacking model. right: cascade forest structure.

### 3.2 Boosting Procedure

Boosting has been a very successful model-guided method to solve two-class or multi-class classification problems [28, 26, 27]. By assigning and updating the weights of each classifier during training, boosting can combine the produced weak learners into a strong classifier [20]. The most important characteristic of boosting is that the process can take advantage of the depth of the proposed deep ensemble model, that is to say, as the number of iterations increases, the distribution of samples is changing, each layer can concentrate more attention to the indistinguishable samples. Therefore, daForest is also an attention-based model. Actually, attention-based models are widely used in various fields, especially in the studies of neural networks [35, 36, 37]. The attention mechanism used in the fields of natural language processing and computer vision is feature-wise attention, whereas, in daForest, we apply sample-wise attention to each layer. There are numerous different implementations of adaptive boosting, such as Adaboost M1/ M2 [25, 26], Adaboost MH [27], SAMME and SAMME.R [28]. As described by Hastie et al. [28], the traditional boosting method is much harder to achieve the minimum requirement for individual learners in the multi-class classification problems, since predictive accuracy must be greater than random guessing accuracy rate 1 / K, where K refers to the number of classes. Accordingly, we choose SAMME.R as the boosting implementation in the proposed model.

As illustrated in Fig. 1, adaptive boosting is used to change the distribution of samples in each layer and draw the attention mechanism into the proposed deep model. Suppose each layer is composed of N individual learners $[F_{l1}, F_{l2}, ..., F_{ln}]$, and we have $\mathcal{L}$ layers, each layer can also be regarded as an individual ensemble module $E_l$. The whole boosting procedure can be described as:

**Algorithm 1** Boosting Procedure

**Require:** original training data $X = [X_0], X_0 \subset \mathbb{R}^d$, class label Y.
1. Initialize the weight of each sample: $w_i^1 = 1/m, i = 1, 2, ..., m$.
2. For $j = 1$ to $\mathcal{L}$:
3.     Fit ensemble module $E_j$ to the training data X:
4.         Fit individual learners $[F_{j1}, F_{j2}, ..., F_{jn}]$ using sample weights.
5.     Obtain the weighted class probability estimates:
6.         $P_k^j(x) = \text{Prob}_w(c = k|x) = E_{jk}(x), k = 1, ..., K$.
7.     Set:
8.         $h_k^j(x) = (K-1)\left(log P_k^j(x) - \frac{1}{K}\sum_{k'} log P_{k'}^j(x)\right), k = 1, ..., K$.
9.     Update training data:
10.         $X = [X_0, h^j(x)], X \subset \mathbb{R}^{d+k*n}$
11.     Set:
12.         $w_i^{j+1} = w_i^j * \exp\left(-\frac{K-1}{K} y_i^T log p^{(j)}(x_i)\right), i = 1, ..., m$
13.     Renormalize $w_i$:
14.         $w_i^{j+1} = w_i^{j+1}/sum(w^{j+1})$
15. Output:
16.     Final-Prediction $= \arg\max_k \sum_j h_k^j(x)$

Cascade forest is transformed to an additive model when the adaptive boosting is applied.

### 3.3 Dense Connectivity

In ordinary cascade forests and gcForest [14], the $(l+1)^{th}$ layer is fed, as described in Sec. 3.1, with the concatenation of $l^{th}$ layer's output and the original training features. We call such structures as sparse connectivity. However, diversity is a crucial metric in ensemble method, each layer of which only has access to the information of its preceding layer in sparse connectivity architecture, leaving much information discarded, which may impede diversity of the ensembles and the information flow in the deep model. We noticed that the training accuracy curve of cascade forest is very unstable after a few rounds of iteration, which may have a noticeable adverse effect on the final performance of the model. Enlightened by the structure of DenseNet [29], we introduce dense connectivity to the ordinary cascade forest and turn it into a dense cascade forest. In our experiments, we find that this structure can effectively suppress the violent concussion of training accuracy curve, making the training process more stable and improving the prediction accuracy to a certain extent.

In dense connectivity, each layer is directly connected to all of the subsequent layers, which can be found in the layout of dense connectivity illustrated by Fig. 1. Consequently, the $l^{th}$ layer receives the probabilistic features of all of the preceding layers. Information received by the $l^{th}$ layer is expressed as:

$$x_l = [x_0, E_{l-1}(x_{l-1}), E_{l-2}(x_{l-2}), \dots, E_1(x_0)], l = 2, \dots, \mathcal{L} \quad (3\text{-}1)$$

$$x_l = E_l(x_0) = [F_{l1}(x_0), \dots, F_{ln}(x_0)], when\ l = 1 \quad (3\text{-}2)$$

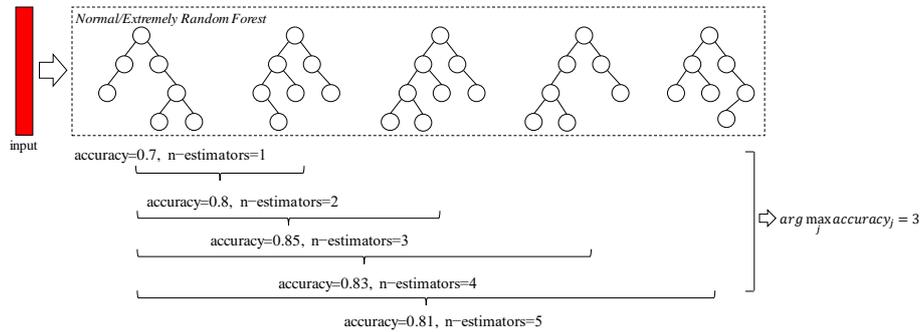

**Fig. 3.** Optimization layer. The layer is a normal random forest or extremely random forest, searching for the optimal number of estimators in a cumulative prediction fashion in all trees. Suppose there are five trees in the optimization layer, the search range will be from 1 to 5 with step size 1.

### 3.4 Hyper-Parameter Optimization Layer

Fig. 3 shows the searching process for the optimal parameters. A forest consists of many estimators (classification and regression trees). The number of estimators is a key hyper-parameter, which has a significant effect on training time and the performance of classification of the deep model, however, it is a time-consuming task to determine the optimal number of estimators by grid-search. In the implementation of gcForest

[14], the number of estimators is fixed to 500, which is not a wise choice since that the complexity of different classification tasks is variant. 500 estimators may be too few or too many for some tasks, leading to underfitting or overfitting respectively. The depth of the models (cascade forest, gcForest and daForest) can be dynamically determined by using the early-stopping mechanism, which will terminate training process if the model does not achieve a continuous improvement on the validation accuracy. Whereas, the quasi-optimal number of estimators in different types of forests must be found in advance. Therefore, we employ a linear search method. Suppose each layer of the deep model is composed of several random forests, to find the quasi-optimal number of estimators in a range of values from 20 to 600 with step size 20, we fit a random forest with 600 estimators and record predictions of each decision tree in the forest on validation dataset before training, then calculate the accuracy on these prediction series. The number corresponding to the highest accuracy is what we are looking for. The whole process to find n-estimators can be described as:

$$\text{nestimators} = \arg\max_j accuracy\left(\frac{\sum_{i=1}^{j} predictions \cdot of \cdot i^{th} \cdot estimator}{j}\right) \quad (4)$$

where $j = 20 + (n-1) * 20, n = 1, \ldots, 30$. The summation term in (4) can be accelerated by utilizing a cache.

## 4  Experiments

We demonstrate the high competitiveness of our model on both high-dimensional sparse datasets and low-dimensional datasets. We compare daForest against other state-of-the-art architectures on each dataset and find that our model has a competitive performance. As described in Sec. 3.1, 3-fold cross validation is applied to generate the probabilistic features in each layer, by which overfitting can be effectively suppressed. We use the original testing set to validate our model when there exists one in the original data set, and if not, we will take 30% of the data set as the testing set and 70% for growing daForest. Ten random runs are adopted to estimate the classification accuracy. We also carry out statistical analysis to validate whether the results obtained by the classification algorithms are significant.

### 4.1  Implementation Details

daForest has one hyper-parameter optimization layer with eight forests in each layer. Initially, each forest in the individual ensemble module has 500 decision trees. If hyper-parameter optimization is activated, the number of decision trees in each forest will be updated according to the results of the linear search as described in Sec.3.4. For some small and low-dimensional datasets (e.g. Yeast, B.C.W.), we will change the search scope to (5, 200) with the step size of 5. Random forests construct each tree by bagging and randomly choosing $\sqrt{d}$ ($d$ represents the dimension of data) features in each inner node split. And we turn an extremely random forest into completely random forest by fixing the maximum number of candidate features in each split to one, which will bring greater diversity to our model, as suggested in [Zhi-Hua Zhou et al. 2017]. In order to abate the impact of abnormal prediction outputs of each layer on the overall results, we introduce the learning rate for boosting procedure and fix it to 0.3, which can also prevent the sample weights from changing too fast. In all experiments, daForest and gcForest are sharing the same cascade structure, as described in Table 1. Other comparative models, such as random forest, logistic regression, SVM and MLP, are trained to the best. The summary of hyper-parameter and default settings are shown in Table 1.

**Table 1.** Configuration of the comparative approaches

| daForest | gcForest | MLP |
|---|---|---|
| **No. hyper-parameter layers:** 1　**Search scope range:** [20, 600] or [5, 200], depending on the scale of each dataset　**Type of forests:** extremely random forest and random forest　**Tree growth:** until all leaves are pure　**No. features for best split:** $\sqrt{d}$ (random forest), 1 (extremely random forest)　**No. Trees in each forest:** dynamically determined by optimization layer　**No. Forests in each layer:** 8　**No. Cascade layers:** 100 iterations | **Type of forests:** extremely random forest and random forest　**Tree growth:** until all leaves are pure　**No. features for best split:** $\sqrt{d}$ (random forest), 1 (extremely random forest)　**No. Trees in each forest:** 500　**No. Forests in each layer:** 8　**No. Cascade Layers:** 100 iterations | **Type of activation function:** ReLu　**No. Hidden layers and layer size:** depending on the scale of each dataset　**Solver for weight optimization:** sgd and adam　**L2 penalty:** {1e-4/ 1e-3}　**Max epochs:** 5000, until convergence　**Batch size:** {32/64/128/256} |

### 4.2 Classification Results on Amazon Commerce Reviews

Amazon Commerce Reviews are derived from customers' reviews in Amazon Commerce Website [38]. As described in Table 2, this data set consists of 1500 samples, where each sample has 10000 attributes. The feature set of A.C.R is composed of four types of writing habits of users: lexical, syntactic, content-specific, and idiosyncratic, each of which has about 1~15 sub-category feature types. There are 19 different feature types in total. The dataset is used in many literatures, but they are all limited to 2~10 reviewing authors. Few classification algorithms extend to a large number of target classes. In order to validate the highly competitive and inspiring robustness achieved by daForest, we compare against ordinary classifiers and deep neural networks on all 50 classes in the dataset.

**Table 2.** Description of Amazon Commerce Reviews

| Language | No. of authors | Reviews per author | Average length of reviews per author | No. of Instance | No. of Attributes |
|---|---|---|---|---|---|
| English | 50 | 30 | 856 characters | 1500 | 10000 |

Table 3 presents the classification accuracy of the proposed algorithm along with other methods. From this table, we can observe that the daForest achieved 84.89% accuracy on American Commence Reviews dataset, which is higher than other competitors, including gcForest by 1.47%. The experimental results demonstrated the superiority of the proposed algorithm when compared with other methods.

In order to compare daForest with gcForest in details, we also perform experiments with different configurations. We typically choose different numbers of layers. The results are reported in Fig. 4. From this figure, we can observe that gcForest outperforms daForest regarding testing accuracy when only one layer is used. However, with the increase of the number of layers, daForest achieves better testing accuracy compared with gcForest, which demonstrates that the proposed algorithm can obtain a more stable result when the model gets deeper.

Features of A.C.R. are not spatial correlated, thus convolutional neural networks cannot be used directly. Therefore, we employ a fully connected deep neural network with 1024-1024-512-256 neurons in each hidden layer. It also includes the experimental results achieved by the synergetic neural network (SNN) [38], which associates the synergetics with artificial neural networks.

**Table 3.** Identification accuracy (%) on Amazon Commerce Reviews

| | | |
|---|---|---|
| | daForest | **84.89 ± 0.26** |
| | gcForest | 83.42 ± 0.61 |
| SNN | Balanced attention parameter | 68.31 [S. Liu, 2011] |
| | Self-adaptive attention parameter | 80.49 [S. Liu, 2011] |
| | SVM (linear kernel) | 60.36 ± 0.46 |
| | SVM (rbf kernel) | 39.83 ± 1.77 |
| | MLP | 60.72 ± 1.91 |
| | Logistic Regression | 72.62 ± 1.70 |
| | Random Forest | 76.60 ± 1.01 |

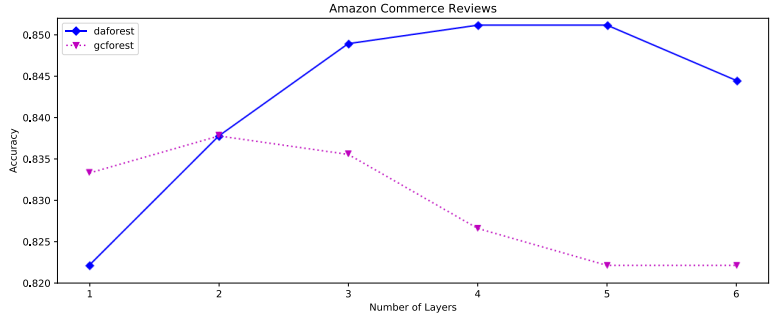

**Fig. 4.** Comparison on testing accuracy between daForest and gcForest

### 4.3 Classification Results on CNAE-9 Data Set

CNAE-9 is a dataset containing 1080 documents of business descriptions of Brazilian companies categorised in 9 categories categorized in the table of National Classification of Economic Activities. The original texts are pre-processed to obtain the current data set: First, we clean up the texts by removing prepositions, punctuations and other tokens. Second, the words are transformed into their canonical forms. Finally, each document is represented as a vector, where the weight of a word is the frequency of it in the document. This data set is highly sparse (99.22% of the matrix is filled with zeros) [41]. The detailed attributes of CNAE-9 are shown in Table 4. We split the dataset into 756 documents for training and 324 documents for testing.

**Table 4.** Description of CNAE-9

| Language | No. of categories | No. of Instance | No. of Attributes |
|---|---|---|---|
| English | 9 | 1080 | 857 |

Since the irrelevance of the dataset, we compare daForest against an MLP classifier with the structure of input-1024-1024-512-256-output. Our experiments also carry out other classification methods and dimensional reduction methods, such as probabilistic neural network based on the evolving system [40], dimensionality reduction based on agglomeration and elimination of terms [39] and SIMACA method coupled with variables selection [41]. As shown in Table 5, daForest shows superior performance over other approaches. Fig. 5 represents the comparison of testing accuracy between daForest and gcForest.

Table 5 presents the classification accuracies of other methods with the proposed algorithm. Observed from this table, daForest achieves an accuracy of 95.81% on the CNAE-9 dataset. Similar with the results in Table 3, it shows that daForest achieves better accuracy than other methods by at least 0.47% and the SVM based methods performed worse than other methods on this dataset, especially rbf kernel-based model which is surpassed by 44.51% by daForest. In terms of classification accuracy, we have improved by 0.90% compared with gcForest. The experimental results demonstrated the superiority of the proposed algorithm in the comparison with other methods.

In order to compare daForest and gcForest in details, we also conducted experiments with different configurations. The results are presented in Fig. 5. From this figure, we can observe that gcForest outperforms daForest in terms of testing accuracy when only few layers are used. However, with the increase of the number of layers, daForest achieves a better testing accuracy compared with gcForest, which demonstrates that the proposed algorithm is able to obtain a more stable result when the model gets deeper.

**Table 5.** Identification accuracy (%) on CNAE-9

| | | |
|---|---|---|
| daForest | | **95.81 ± 0.20** |
| gcForest | | 94.91 ± 0.42 |
| ePNN | ePNN1 | 88.71% [P. Marques, 2010] |
| ePNN | ePNN2 | 84.45% [P. Marques, 2010] |
| Dim. Reduction | MI_1 | 92.78% [P. Marques, 2009] |
| Dim. Reduction | IAE | 91.11% [P. Marques, 2009] |
| SIMCA | VSC-SIMCA | 95.34% [Saleh, 2015] |
| SIMCA | SIMCA | 94.62% [Saleh, 2015] |
| SVM (linear kernel) | | 93.02 ± 0.16 |
| SVM (rbf kernel) | | 51.30 ± 6.76 |
| MLP | | 95.01 ± 0.34 |
| Logistic Regression | | 94.41 ± 0.80 |
| Random Forest | | 89.72 ± 0.57 |

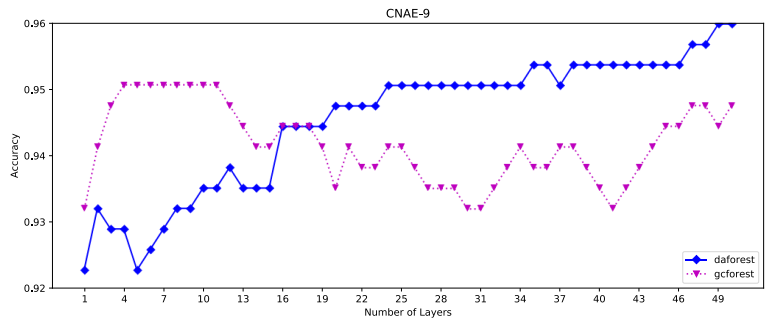

**Fig. 5.** Comparison of testing accuracy between daForest and gcForest

### 4.4 Classification Results on IMDB Movie Reviews

The IMDB movie reviews dataset [43] contains 50000 reviews, each of which is labeled by sentiment (positive or negative). We split the dataset into 25000 for training and 25000 for testing. Reviews have been processed and encoded as sequences of indexes, each feature of the sequences is a word index, which can be regarded as the ranking of overall frequency of one word in the IMDB dataset. For example, integer 1 indicates the most frequent word in the dataset. We select the 5000 most frequent words as training and testing attributes. To adjust the data to our model, we apply tf-idf transformation to the reviews. So, the attributes of samples are represented as the product of term-frequency and inverse document-frequency of corresponding words. Table 6 represents the details of IMDB.

**Table 6.** Description of IMDB

| Language | No. of categories | No. of Instance | No. of Attributes |
|---|---|---|---|
| English | 2 | 50,000 | 5,000 |

We compare daForest with CNNs trained on word vectors [14, 44] and an MLP with hidden layers of shape 1024-1024-512-256. It also includes the results of other traditional classification models, as shown in Table 7. And Fig. 6 represents the comparison on testing accuracy between daForest and gcForest. gcForest is automatically terminated after few iterations when accuracy stops increasing.

The classification accuracy of the proposed approach and other methods are reported in Table 7. From this table, we can observe that daForest achieves an accuracy of 89.37% on IMDB dataset. Same to previous results, it is the highest among all methods. Besides, compared with gcForest, daForest has improved 0.25% in terms of classification accuracy.

We also consider applying different numbers of layers to the comparison between daForest and gcForest. The detailed results are presented in Fig. 6, where we can find that the two methods perform similarly but daForest catches up with gcForest on accuracy as the number of layers surpasses 16. It is demonstrated that the proposed algorithm is able to obtain a more accurate and more stable prediction with the increase of the number of layers.

**Table 7.** Identification accuracy (%) on IMDB

| | |
|---|---|
| daForest | **89.37 ± 0.16** |
| gcForest | 89.12 ± 0.17 |
| CNN | 89.02% [Zhou, 2017] |
| SVM (linear kernel) | 88.13 ± 0.26 |
| SVM (rbf kernel) | 54.92 ± 0.28 |
| MLP | 85.80 ± 0.48 |
| Logistic Regression | 88.60 ± 0.05 |
| Random Forest | 85.07 ± 0.08 |

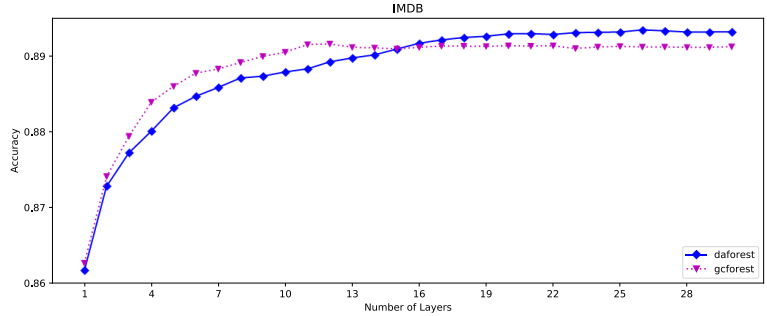

**Fig. 6.** Comparison on testing accuracy between daForest and gcForest

### 4.5 Classification Results on Low-Dimensional Datasets

In this section, we validate our model on 5 UCI datasets [42]: Letter Dataset, Adult Dataset, Yeast Dataset, Breast Cancer Wisconsin (Original) Dataset and Parkinsons Dataset. Letter contains 16000/4000 instances for training/testing. Adult contains 34190/14652 for training/testing. Yeast contains 1039/445 for training/testing. B.C.W contains 490/209 for training/testing. Parkinsons contains 138/59 for training/testing. Other details of the datasets are shown in Table 8. We employ different MLP configurations on each dataset in our experiment as [Zhou, 2014] suggested. CNNs are not applicable on these low-dimensional datasets, especially for whose datasets which have instances of less than 2,000. Classification results are described in Table 9. It is noteworthy that SVM on Adult is unable to converge automatically, so we limit the max iterations of SVM to 1000.

The classification accuracies of various low dimensional datasets are summarized in Table 8. From this table, we can observe that the proposed algorithm achieves the best performance on all low dimensional datasets when compared with other methods. It demonstrates that the proposed algorithm not only performs better on those complicated datasets, but also has superiority on low dimensional datasets.

**Table 8.** Details of UCI datasets

| Dataset name | No. of categories | No. of Instance | No. of Attributes |
|---|---|---|---|
| Letter | 26 | 20000 | 16 |
| Adult | 2 | 48842 | 14 |
| Yeast | 10 | 1484 | 8 |
| B.C.W | 2 | 699 | 10 |
| Parkinsons | 2 | 197 | 23 |

**Table 9.** Identification accuracy (%) on UCI datasets

| Method | LETTER | ADULT | YEAST | B.C.W. | Parkinsons |
|---|---|---|---|---|---|
| daForest | **98.06 ± 0.18** | **86.20 ± 0.05** | **64.88 ± 0.25** | **97.62 ± 0.00** | **87.01 ± 1.07** |
| gcForest | 97.34 ± 0.09 | 86.02 ± 0.12 | 63.30 ± 0.80 | **97.62 ± 0.00** | 84.32 ± 1.43 |
| SVM (linear kernel) | 86.58 ± 0.08 | 76.38 ± 0.00 | 56.88 ± 0.26 | 96.62 ± 0.15 | 78.65 ± 0.87 |
| SVM (rbf kernel) | 97.88 ± 0.04 | 76.38 ± 0.00 | 56.55 ± 0.51 | 96.62 ± 0.15 | 82.04 ± 1.18 |
| MLP | 95.51 ± 1.02 | 85.30 ± 0.62 | 55.13 ± 1.36 | 95.76 ± 0.61 | 74.00 ± 4.45 |
| Logistic Regression | 72.27 ± 0.47 | 79.81 ± 0.20 | 52.60 ± 0.89 | 96.67 ± 0.39 | 81.02 ± 2.74 |
| Random Forest | 97.04 ± 0.34 | 85.11 ± 0.05 | 61.69 ± 0.40 | 97.14 ± 0.00 | 84.41 ± 1.56 |

## 4.6 Significance Test

We perform Friedman's test and Iman-Davenport's test on the results obtained by the classification algorithms, shown in Table 10, to verify whether there are significant differences. $\alpha = 0.05$ is used as the level of confidence in all algorithms. Once we confirm that there are significant differences in these results, we can carry out the Wilcoxon testing as a Post-Hoc test to compare the best performing classifier against other algorithms. Table 11 represents the results of Friedman's test and Iman-Davenport's test based on the accuracies of classifiers. Results of the Wilcoxon testing on accuracy are shown in Table 12.

From Table 11 and Table 12, we can observe that the proposed algorithm has the lowest p-value, which demonstrates the strong competitiveness of the proposed algorithm in terms of probability and statistics.

**Table 10.** Experiment results (mean accuracy) of the proposed model and other classifiers

| Dataset | daForest | gcForest | SVM L. | SVM R. | MLP | L.R. | R.F. |
|---|---|---|---|---|---|---|---|
| A.C.R. | 84.89% | 83.42% | 60.36% | 39.83% | 60.72% | 72.62% | 76.60% |
| CNAE-9 | 95.81% | 94.91% | 93.02% | 51.30% | 95.01% | 94.41% | 89.72% |
| IMDB | 89.37% | 89.12% | 88.13% | 54.92% | 85.80% | 88.60% | 85.07% |
| LETTER | 98.06% | 97.34% | 86.58% | 97.88% | 95.51% | 72.27% | 97.04% |
| ADULT | 86.20% | 86.02% | 76.38% | 76.38% | 85.30% | 79.81% | 85.11% |
| YEAST | 64.88% | 63.30% | 56.88% | 56.55% | 55.13% | 52.60% | 61.69% |
| B.C.W. | 97.62% | 97.62% | 96.62% | 96.62% | 95.76% | 96.67% | 97.14% |
| Parkinsons | 87.01% | 84.32% | 78.65% | 82.04% | 74.00% | 81.02% | 84.41% |
| Arrhythmia | 73.72% | 72.92% | 66.49% | 54.16% | 67.79% | 62.77% | 71.90% |
| Car | 98.25% | 97.76% | 93.07% | 94.63% | 99.24% | 87.18% | 96.07% |
| Credit | 89.08% | 88.79% | 54.37% | 55.52% | 65.99% | 86.63% | 88.74% |
| Contraceptive | 54.78% | 53.68% | 48.73% | 55.77% | 51.83% | 49.64% | 51.29% |
| German | 77.82% | 77.01% | 75.60% | 71.20% | 75.43% | 76.60% | 76.73% |
| Glass | 86.38% | 85.12% | 66.15% | 73.54% | 59.63% | 61.32% | 83.69% |

**Table 11.** Results of Friedman's and Iman–Davenport's tests on the accuracies

| Method | Statistical value | p-Value | Hypothesis |
|---|---|---|---|
| Friedman | 50.716837 | <0.005 | Rejected |
| Iman–Davenport | 19.809381 | <0.005 | Rejected |

**Table 12.** Results of the Wilcoxon testing on the accuracies (daForest is the control algorithm)

| i | Algorithm | p-Value | Hypothesis |
|---|---|---|---|
| 1 | gcForest | 0.001097 | Rejected |
| 2 | SVM (linear kernel) | 0.000982 | Rejected |
| 3 | SVM (rbf kernel) | 0.001523 | Rejected |
| 4 | MLP | 0.001887 | Rejected |
| 5 | Logistic Regression | 0.000982 | Rejected |
| 6 | Random Forest | 0.000982 | Rejected |

### 4.7 Running Time

Our experiments are carried out on a personal computer with an AMD Ryzen1400 CPU (4 cores) and 16G RAM. In the experiments of this section, we disable early-stopping mechanism and force it to predict probabilistic features for both training and testing samples after each epoch, so the time consumed is actually much higher than a pure training process. Even so, the time expenditures of daForest are still acceptable: On Amazon Commerce Reviews, daForest converges after 4 epochs (layers) with about 93 seconds taken per epoch. On CNAE-9 dataset, it converges after 50 epochs with about 19.2 seconds taken per epoch. It takes 25 epochs to converge on IMDB dataset with about 15 minutes taken for training and prediction processes, however, it is noteworthy that gcForest's cost in training and prediction can be reduced to 4 minutes per epoch on a PC with 2 Intel E5 2695 v4 CPUs as reported in [Zhou, 2017] on IMDB, and time expenditures of daForest and gcForest are quite the same according to the experiments. As is known, a random forest is an inherently parallel learning model. And daForest, which is made up of random forest and extremely random forest, also has a very good level of parallel computation, leading to a high probability of considerable reduction on training time in the further improvements of daForest. The detailed results on the running time of the proposed method and gcForest are shown in Table 13, with minute being the unit of time. To compare the proposed model and gcForest more detailly and make it comparable on time consumption, we compared both models on the same number of cascade layer, which is making the time cost comparison on the layer where the search process goes to a stable point. The inner structures of the layers are the same, making da/gc-forest both have 8 forests, but for daForest, the number of trees in each forest is dynamically determined by the optimization layer, hence, in some cases, daForest has slightly less consumption of time.

Table 13. The running time (minutes) of daForest and gcForest on each dataset

| Cost | daForest | | gcForest | |
|---|---|---|---|---|
| | total | layers | total | layers |
| A.C.R. | 6.16 | 4 | 5.40 | 4 |
| CANE-9 | 16.43 | 50 | 17.25 | 50 |
| LETTER | 15.50 | 6 | 14.20 | 6 |
| ADULT | 10.38 | 6 | 13.39 | 6 |
| YEAST | 2.45 | 3 | 2.36 | 3 |
| B.C.W. | 2.00 | 5 | 1.70 | 5 |
| Parkinsons | 3.08 | 5 | 2.63 | 5 |
| IMDB | 367.50 | 25 | 340.75 | 25 |

## 5 Conclusion

It is widely recognized that deep model has achieved great success in exploiting high-order features. The most popular deep models are still deep neural networks, which are the killer architectures in image and audio recognition domains. However, in many fields, the most frequently used datasets are still small-scale or space/time uncorrelated datasets, where deep neural networks cannot be fully applied. It is still interesting and necessary to explore other deep architectures. gcForest is one of such pioneering works with great creativity and achieves highly competitive performance compared with traditional learning models and deep neural networks. However, gcForest is not a perfect model since degeneration limits the expression ability and depth of the model and the accuracy drops in few epochs. It is very hard to push the model to go deeper. In this paper, we propose a new deep ensemble model, which we refer to as Dense Adaptive Cascade Forest (daForest). By employing boosting procedure and dense connectivity, the proposed model remedies the degeneration to a considerable extent and tent to yield consistent accuracy improvement as the number of layers increases. daForest achieves state-of-the-art results across several highly competitive datasets. Moreover, daForest has less training and prediction time when the optimization layer is activated, which is very crucial when the training set is large and running time is crucial.

There are several directions for future works. First, we can adjust daForest for the task of image recognition. For example, we can add several convolutional layers to extract high-order features and replace the fully connected layer with daForest. Second, ordinary decision trees use hard splits in each internal node, which is undifferentiable, so we need to train the whole model layer by layer and push it to go deeper by employing boosting procedure and dense connectivity. However, layer by layer training is not an effective method to construct a deep model, so we might to use soft splits in each decision tree to achieve end-to-end training with daForest in future work.

## Acknowledgement

This research was sponsored by National Key R&D Plan of China (No. 2017YFB1400300 and No. 2017YFB1400303).

**Compliance with ethical standards**

**Conflict of interest** The authors declare that there is no conflict of interests regarding the publication of this paper.

**Ethical approval** This article does not contain any studies with human participants or animals performed by any of the authors.